\documentclass[11pt]{article}

\usepackage[final]{acl}

 \usepackage{microtype}

\usepackage[english,bidi=default]{babel} 
\babelfont{rm}{TeXGyreTermesX} 
\babelprovide[import]{hindi}
\babelfont[*devanagari]{rm}{Lohit Devanagari}
\babelprovide[import]{arabic}
\babelfont[*arabic]{rm}{Noto Sans Arabic}

\usepackage{polyglossia}
\setdefaultlanguage{english}
\setotherlanguages{arabic,russian,thai,hindi,kannada}

\usepackage{graphicx}
\usepackage{booktabs}

\usepackage{amsmath}
\usepackage{annotate-equations}
\usepackage{tikz}
\usepackage{xcolor}
\usepackage{hyperref}
\usepackage{todonotes}
\usepackage{multirow}
\usepackage{subcaption}
\usepackage{kotex}
\usepackage{soul}

\definecolor{green}{HTML}{DAEEE0}
\definecolor{pink}{HTML}{F0DEE7}
\definecolor{gray}{HTML}{EDECE7}

\newcommand{\emoji}[1]{%
  \begingroup\normalfont\Large\raisebox{-4pt}{\includegraphics[height=12pt]{#1}}\endgroup
}

\title{The Multilingual FrameNet Corpus}

\author{
    \textbf{Beatrice Fiumanò\textsuperscript{1,*}},
    \textbf{Nicolas Lazzari\textsuperscript{1,2}},
    \textbf{Simone Paolo Ponzetto\textsuperscript{3}},
    \textbf{Valentina Presutti\textsuperscript{1}}
    \\
    \\
    \textsuperscript{1}University of Bologna, Italy,
    \textsuperscript{2}University of Pisa, Italy, \\
    \textsuperscript{3}University of Mannheim, Germany \\
    \small{\texttt{\{beatrice.fiumano,nicolas.lazzari3,valentina.presutti\}@unibo.it}}, \ \small{\texttt{ponzetto@uni-mannheim.de}} \\
    \small{\textsuperscript{*}\textbf{Corresponding author}}
}

\begin{document}
\maketitle
\begin{abstract}
This paper introduces the Multilingual FrameNet Corpus (mFNC), a novel resource that extends the English Berkeley FrameNet corpus by collecting and harmonizing existing language-specific corpora across nine additional languages: Brazilian Portuguese, Chinese, Dutch, French, German, Italian, Korean, Latvian and Swedish. 

By training models that rely on different architectures on the mFNC, we consistently outperform existing state-of-the-art Frame Semantic Parsers in both multilingual and cross-lingual settings, underscoring the importance of multilingual training data. 

The mFNC and our trained FSP models are openly available at \url{https://github.com/beatrice-f/mFNC}.
\end{abstract}

\section{Introduction}
\label{sec:intro}

Frame Semantic Parsing (FSP) is the task of automatically identifying semantic frames in text according to Fillmore's Frame Semantics theory \cite{fillmore1976frame}. FSP has proven beneficial across several NLP tasks, such as information extraction \citep{li-etal-2025-frame}, Knowledge Graph construction \citep{alam2021srlkg}, opinion mining \cite{RecuperoPCGN15}, sentiment analysis \cite{AtzeniDR18} and framing detection \cite{minnema-etal-2022-sociofillmore, COSCHIGNANO2023107}.

\begin{figure}[t!]
    \centering
    \includegraphics[width=\linewidth]{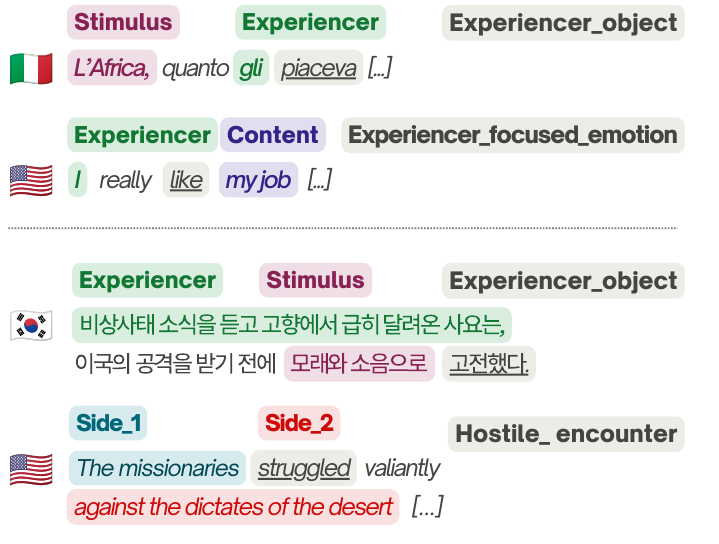}
    \caption{\textbf{Example of annotated sentences from the mFNC.} Lexical Units are underlined. Textual spans classified by a Frame Elements are shaded using the same color. \emoji{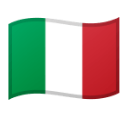}: ``\textit{\sethlcolor{pink}\hl{Africa}, how much \sethlcolor{green}\hl{he} \sethlcolor{gray}\hl{liked} it [...]}''. \emoji{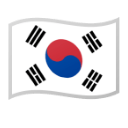}: \textit{``\sethlcolor{green}\hl{The person who rushed back from their hometown upon hearing news of the emergency} \sethlcolor{gray}\hl{struggled} \sethlcolor{pink}\hl{with sand and noise} before facing attack from a foreign country.''.}}
    \label{fig:corpus_example}
\end{figure}

Despite its wide range of applications, the multilingual dimension of FSP remains largely underexplored. Recent state-of-the-art (SotA) approaches (e.g., \citet{devasier-etal-2024-robust}) are trained and evaluated exclusively on the Berkeley FrameNet (BFN) corpus \citep{Baker98}, with isolated attempts at cross-lingual transfer through multilingual Transformer-based architectures (e.g., \citet{xia-etal-2021-lome}). Unlike other tasks, however, FSP presents unique challenges due to cross-lingual variations in the grammatical, cultural and conceptual dimensions of language.
For instance, in the Italian ``piacere'' (to like), the thing being liked acts upon the liker, while in English it is the liker who actively likes something \citep{baker-lorenzi-2020-exploring}. Similarly, the Korean expression ``고전하다'' (to struggle) casts the struggling entity as acted upon by an external force, while English frames the same situation with the struggling entity as an active participant. These examples illustrate how languages can conceptualize the same situation in different ways, expressing it through distinct syntactic and semantic patterns. As shown in Figure \ref{fig:corpus_example}, these differences are captured by Frame Semantics.


The lack of a unified large-scale multilingual corpus annotated using BFN is largely responsible for this under-exploration, preventing FSP systems from being trained or tested on languages beyond English. Although numerous independent language-specific FrameNets have been developed since the inception of BFN, these resources remain scattered and heterogeneous in their format and annotation scheme, hampering integration. Approaches to automatically align them have been proposed \citep{baker-lorenzi-2020-exploring}, but to the best of our knowledge no unified corpus exists to date.

To address this gap, we present the mFNC (Multilingual FrameNet Corpus), a novel multilingual dataset that collects and harmonizes ten language-specific corpora annotated using BFN. Alongside English, the mFNC extends coverage to Brazilian Portuguese, Chinese, Dutch, French, German, Italian, Korean, Latvian and Swedish. Figure \ref{fig:corpus_example} shows an example of sentences and annotations from the mFNC.

Using the mFNC, we demonstrate that FSP models trained on multilingual data consistently outperform models trained on English only. Beyond FSP, the mFNC contributes to the broader goal of a unified Multilingual FrameNet \citep{GILARDI18.11}, offering an empirical basis to investigate how frame-semantic structures vary across typologically diverse languages \citep{ellsworth-etal-2021-framenet}.

In summary, our contribution is two-fold:
\begin{enumerate}
    \item We present the mFNC, a multilingual benchmark that collects and harmonizes \textbf{ten} language-specific FrameNets;
    \item We openly share three multilingual FSP systems that outperform existing SotA baselines in multilingual and cross-lingual settings.
\end{enumerate}

The rest of the paper is structured as follows: Section \ref{sec:rw} introduces the background on Frame Semantics and reviews existing work on BFN applications, resources that extend BFN to other languages, and existing FSP approaches. Section \ref{sec:dataset} describes the resources used to construct the mFNC and provides insights on them. Section \ref{sec:experiments} presents a comparative evaluation of SotA FSP models trained on BFN and on the mFNC, in both multilingual and cross-lingual settings. Section \ref{sec:discussion} discusses our findings and outlines directions for future work. Finally, Section \ref{sec:conclusion} summarizes our contributions and Section \ref{sec:limitations} highlights limitations.

\section{Related Work}
\label{sec:rw}


Before reviewing multilingual FrameNet resources and parsers, we provide an example of how a natural language sentence is annotated using BFN. Consider the sentence ``I really like my job'', shown in Figure \ref{fig:corpus_example}.

The \textit{Lexical Unit} (LU) ``like'' \textit{evokes} the frame \textsc{Experiencer\_focused\_emotion}, which represents the concept of ``someone experiencing some emotion with respect to some content''. In turn, the span ``I'' is classified by the \textit{Frame Element} (FE) \textsc{Experiencer}, while ``my job'' is classified by the FE \textsc{Content}. 

BFN defines a large set of frames, FEs, and English LUs. For a complete treatment, we refer the reader to \citet{ruppenhofer2016framenet}.


\subsection{Toward a Multilingual FrameNet}
While BFN originally focused on English, extending it to a multilingual setting has been a central objective of the project since its inception. This ambition has driven the development of numerous language-specific resources, and resulted in the Global FrameNet project\footnote{\url{https://www.globalframenet.org/}}, which aims both at aligning existing datasets \cite{baker-lorenzi-2020-exploring} and at creating new multilingual ones through shared annotation tasks \cite{GILARDI18.11}.

As with other lexical-semantic networks, the creation of new FrameNets follows distinct lexicographic strategies, such as projecting the BFN frame inventory onto another language, developing a new culture-specific inventory, or combining both practices by reusing and expanding the BFN repertoire \cite{Boas-multilingualFNs}. As detailed in Section \ref{sec:dataset}, we focus on resources that fully or partially reuse the English set of frames, as this enables cross-lingual alignment and the construction of a cohesive dataset.

\subsection{Frame Semantic Parsing}

The FSP task consists of automatically annotating a natural language sentence with semantic frames, as shown in Figure \ref{fig:corpus_example}. The task is usually decomposed into four sequential sub-tasks: detecting the frame-evoking LUs, classifying them with the correct frame, and identifying the spans classified by the FEs of the frame. Existing FSP systems either address each step in isolation or jointly solve multiple sub-tasks. 

\paragraph{FSP Applications}
As introduced above, FSP enables structuring natural language text according to semantic frames and FEs, making its predicate-argument structure explicit.

This approach has proven useful across a wide range of NLP tasks. Alongside the applications listed in Section \ref{sec:intro}, \citet{Taniguchi-legalQA} use FSP for yes/no QA to extract predicate-argument configurations from legal texts, enabling semantic matching between questions and candidate answers beyond surface string overlap. More recently, \citet{li-etal-2025-frame} proposed FrameRTE, a three-stage pipeline that combines an FSP's output and LLMs for zero-shot relation triplet extraction. In text summarization, frame-based graphs have been exploited to improve salience estimation for both extractive \cite{guan-etal-2021-frame} and abstractive \cite{GUAN2021106973} summarization. Beyond core NLP tasks, FSP has also been leveraged for applications at the discourse level. Frame-based embeddings have been used to train transformer models for metaphor detection \cite{li-etal-2023-framebert} and generation \cite{stowe-etal-2021-metaphor}. \citet{remijnse-etal-2024-tracking} use frames to analyze framing and perspectivization of events' participants across documents, while \citet{ryazanov2025chatgpt} and \citet{fiuman-etal-2026-victim} apply FSP to trace differences in media narratives. In light of these applications, the utility of FSP clearly extends across all languages, reinforcing the need for multilingual systems.

\paragraph{Multilingual FSP} 
In general, SotA models are trained and tested on the BFN corpus \cite{swayamdipta:17, das-etal-2010-probabilistic, kalyanpur2020frametransformer, devasier-etal-2024-robust}, while models operating in multilingual or cross-lingual settings have received less attention. Notable exceptions include \citet{johannsen-etal-2015-language}, where the model by \citet{das-etal-2010-probabilistic} is extended with multilingual word embeddings and its cross-lingual performance is evaluated on a novel corpus of nine languages. Although the corpus is too small to train an FSP from scratch, the results demonstrate that a multilingual backbone is beneficial for cross-lingual generalization.

More recently, \citet{xia-etal-2021-lome} build upon this finding by proposing LOME, an FSP model that fine-tunes the multilingual XLM-RoBERTa encoder model on the BFN corpus. However, the model is only evaluated on the English language. 

To the best of our knowledge, no FSP models trained and evaluated in a multilingual setting have been proposed to date. 

\paragraph{LLM-driven FSP} 
Beyond specialized approaches to FSP, LLMs represent valid language-agnostic tools thanks to their broad multi-language coverage. Unlike previous approaches, their application so far has been limited to individual FSP sub-tasks, such as frame identification \cite{Chundru25} and FE classification \cite{DevasierLLMsArguments}, or in modular pipelines \cite{yahui-etal-2024-leveraging}. Results have shown that LLMs underperform in zero- and few-shot settings, narrowing the performance gap with specialized FSP models only through fine-tuning.



\section{mFNC: \underline{M}ultilingual \underline{F}rame\underline{N}et \underline{C}orpus}
\begin{table*}[ht]
    \centering
    \begin{tabular}{cccccrrrr}    
    \toprule
    & & Tokens & \multicolumn{1}{c}{Frames} & \multicolumn{1}{c}{FEs} & \multicolumn{4}{c}{\# Sentences}  \\ \cmidrule{6-9} 
    & & & & & Train & Valid. & Test & Total \\
    \midrule
    DE & \emoji{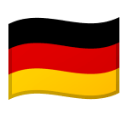} & 345247 & 19688 (230) & 35597 (848) & 10543 & 1629 & 2989 & 15161 \\
    EN & \emoji{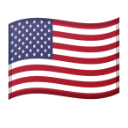} & 112907 & 25951 (788) & 45639 (3668) & 3312 & 325 & 1211 & 4848 \\
    FR & \emoji{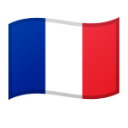} & 206729 & 6536 (51) & 12514 (198) & 3733 & 558 & 1043 & 5334 \\
    IT & \emoji{emoji-imgs/it.png} & 25585 & 1255 (192) & 2860 (726) & 621 & 105 & 225 & 951 \\
    KO & \emoji{emoji-imgs/kr.png} & 153391 & 21094 (800) & 38972 (4226) & 8407 & 1395 & 2188 & 11990 \\
    LV & \emoji{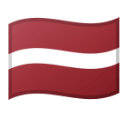} & 299341 & 13527 (242) & 24836 (1131) & 9451 & 1350 & 2726 & 13527 \\
    NL & \emoji{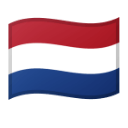} & 21128 & 1339 (200) & 2054 (479) & 713 & 104 & 218 & 1035 \\
    PT & \emoji{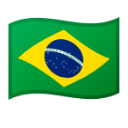} & 28236 & 7767 (519) & 12685 (1730) & 1319 & 171 & 433 & 1923 \\
    SV & \emoji{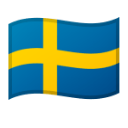} & 122939 & 8018 (954) & 17209 (5174) & 5506 & 769 & 1694 & 7969 \\
    ZH & \emoji{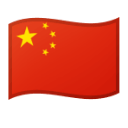} & 189257 & 9107 (584) & 23202 (3660) & 4365 & 786 & 1347 & 6498 \\
    total & \emoji{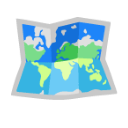} & 1504760 & 114282 (1070) & 215568 (8954) & 47970 & 7192 & 14074 & 69236 \\
    \bottomrule
    \end{tabular}
    \caption{Composition of the mFNC. Unique frames and FEs are reported within the parenthesis.}
    \label{tab:dataset-composition-full}
\end{table*}

In this section we present the mFNC which, to the best of our knowledge, is the first large-scale multilingual corpus annotated using BFN. We construct the mFNC by collecting ten different corpora annotated using BFN and harmonizing them in a shared format. In the next sections we first describe our data selection (Section \ref{sec:data-selection}) and harmonization (Section \ref{sec:data-harmonization}) processes, and later provide insights on the mFNC statistics (Section \ref{sec:dataset-composition}) and content (Section \ref{sec:content-composition}).

\subsection{Data Selection}
\label{sec:data-selection}
\label{sec:dataset}
\begin{table}[ht]
    \centering
    \begin{tabular}{cccc}
    \toprule
    \multicolumn{2}{c}{\textbf{Lang.}} & \textbf{Source} & \textbf{Acc.} \\
    \midrule
    DE & \emoji{emoji-imgs/de.png} & \citep{rehbein2012salsa} & \href{https://www.coli.uni-saarland.de/projects/salsa/}{UR} \\
    EN & \emoji{emoji-imgs/us.png} & \citep{Baker98} & \href{https://www.nltk.org/howto/framenet.html}{OA} \\
    FR & \emoji{emoji-imgs/fr.png} & \citep{djemaa2016asfalda} & \href{https://github.com/kleag/french-framenet/}{OA} \\
    IT & \emoji{emoji-imgs/it.png} & \citep{venturi2009iframe} & \href{http://sag.art.uniroma2.it/iframe/doku.php?id=resources:pisa:manual_salsa_style}{OA} \\
    KO & \emoji{emoji-imgs/kr.png} & \citep{kim-etal-2016-korean} & \href{https://github.com/machinereading/koreanframenet}{OA} \\
    LV & \emoji{emoji-imgs/lv.png} & \citep{gruzitis2018latvian} & \href{https://github.com/LUMII-AILab/FullStack}{OA} \\
    NL & \emoji{emoji-imgs/nl.png} & \citep{vossen-etal-2020-large} & \href{https://web.archive.org/web/20260112195820/https://www.dutchframenet.nl/data-releases/}{OA} \\
    PT & \emoji{emoji-imgs/br.png} & \citep{belcavello-etal-2024-frame2} & \href{https://github.com/FrameNetBrasil/frame-squared}{OA} \\
    SV & \emoji{emoji-imgs/se.png} & \citep{dannells2021swefn} & \href{https://spraakbanken.gu.se/en/resources/swefn}{OA} \\ 
    ZH & \emoji{emoji-imgs/cn.png} & \citep{you2005chfn} & \href{https://tianchi.aliyun.com/dataset/149079}{UR} \\
    \bottomrule
    \end{tabular}
    \caption{FrameNet resources selected to construct the mFNC. Datasets marked with \textit{OA} are openly accessible, while datasets marked with \textit{UR} are available upon request.}
    \label{tab:datasets}
\end{table}

Table \ref{tab:datasets} provides an overview of the ten original resources used to construct the mFNC.

The resources vary in their development strategy and the type of language data they annotate. However, they converge in their full or partial adoption of BFN frames and FEs. 

\paragraph{Bottom-up approaches}
With the exception of Swedish and Korean, all resources adopt a bottom-up approach, meaning they derive frame annotations from existing or newly collected language-specific corpora and treebanks, with varying degrees of adherence to BFN. Although more labor intensive, this method preserves the conceptual structure of the source language and avoids being constrained by the set of BFN's LUs. 

\paragraph{Top-down approaches}
Korean FrameNet is instead developed using a top-down (or extension \cite{dannells2021swefn}) approach, where sentences from the BFN corpus are translated into Korean, enabling the identification of language-specific LUs. While this approach is more efficient, it forces an English-centric conceptualization on the new resource. At a later stage, the Korean FN was expanded via cross-language projection of the Japanese FrameNet\footnote{The resource is currently not publicly available, and we were not able to access it.}\cite{ohara2004japanese}. 

\paragraph{Hybrid construction}
The Swedish FrameNet is developed following a hybrid approach, first reusing BFN frames and translating English LUs into Swedish, and later defining new frames and LUs by adopting a corpus-based approach.

\paragraph{Document genre}
The ten resources also differ in the genre of annotated documents, spanning newspaper articles (German, Italian, partially French), multi-genre treebanks covering medical, legal, and parliamentary texts (French), large general-domain and specialized corpora (Chinese), multi-genre texts reporting on selected event types (Dutch), a mixed corpus of news, fiction, legal, and spoken texts (Latvian), and informal sources such as TV series transcripts (Brazilian Portuguese). This diversity contributes to the richness of topics and styles in the mFNC documents.

\subsection{Data Harmonization}
\label{sec:data-harmonization}
We seek to construct the mFNC to contain both plain and tokenized documents, aligning with the original BFN corpus format. However, the ten FrameNets differ in how they provide the documents, sometimes offering only the plain or tokenized text. 

To achieve a fully harmonized resource, we recover missing full-text documents from their tokenized version using the language-specific detokenizers available in the SacreMoses Python library\footnote{\url{https://github.com/hplt-project/sacremoses}}. Similarly, we tokenize full-text documents that are not already tokenized using the same library.

Finally, we collect all the annotations for each document. An annotation is defined by the LU that evokes a frame, the frame, and the list of annotated FEs. We remove language-specific frames for those resources that follow a hybrid annotation approach, to ensure full cross-language compatibility with BFN.\footnote{In the GitHub repository, we also provide a complementary version of the mFNC that preserves language-specific annotations.}

\subsection{mFNC composition}
\label{sec:dataset-composition}
Table \ref{tab:dataset-composition-full} reports an overview of the composition of the mFNC after the harmonization phase. In total, the mFNC includes 1.5M tokens collected from approximately 70k sentences, accumulating a total of over 100k annotated frames and 200k annotated FEs. To support reproducibility, for English we re-use the splits already available for the BFN corpus, as computed by \citet{swayamdipta:17}\footnote{We only include the FrameNet 1.7 corpus as it encompasses the previous 1.5 version.}. Following the same work, we only retain annotations of the full-text documents, discarding annotations of the exemplar sentences provided for each frame. Indeed, \citet{das-etal-2014-frame} noted that training on exemplar sentences hurt model performance, probably due to their lack of representativeness and incomplete annotations. 
We compute novel splits for the remaining languages by prioritizing a balanced distribution of frames \citep{sechidis2011stratification}.

\paragraph{Frames coverage}
\begin{figure}[htbp]
    \centering
    \includegraphics[width=\linewidth]{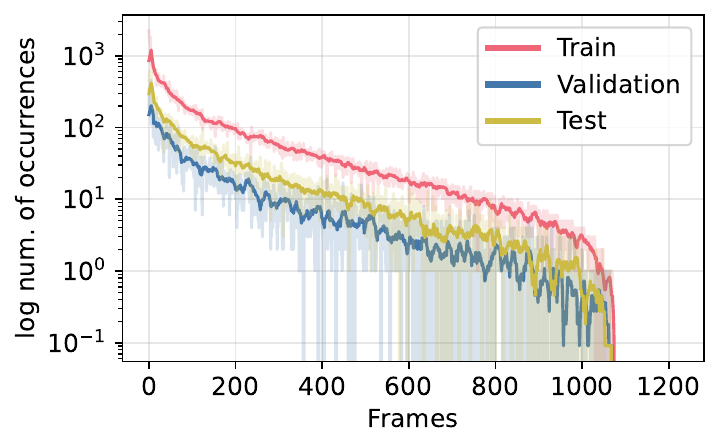}
    \caption{Number of occurrences of \textit{all} BFN frames (in log space) in each mFNC split.}
    \label{fig:splits-occurences}
\end{figure}
Figure \ref{fig:splits-occurences} shows that the number of occurrences of each frame follows a Zipfian distribution, hinting at its highly unbalanced nature inherited from the resources of Table \ref{tab:datasets} (see also Figure \ref{app:fig:splits-occurrences-per-language} in the Appendix for language-specific breakdown). 

For example, the frames \textsc{Causation} and \textsc{Statement} are the two most frequently annotated frames, while \textsc{Causation\_scenario} and \textsc{Explosion} are both annotated only once in the mFNC (see Table \ref{app:tab:top-5} in the Appendix for the most common frames annotated for each language). We report that 151 of the 1221 total frames in BFN ($\approx 12\%$) never occur in the mFNC annotations. 

\paragraph{Annotation density}
\begin{figure}[ht]
    \centering
    \includegraphics[width=\linewidth]{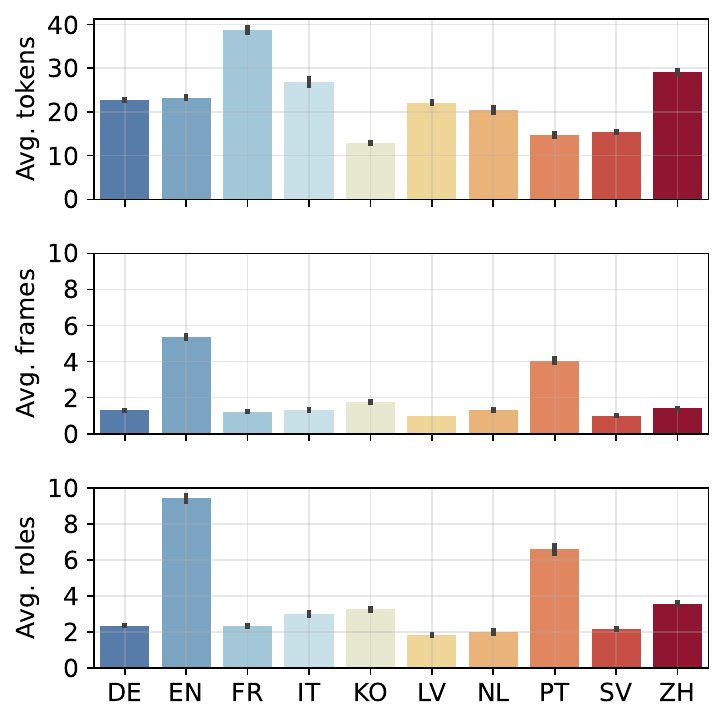}
    \caption{Average number of tokens, frames and FEs annotated in each document.}
    \label{fig:sentence-stats}
\end{figure}
The resources differ in the density of available annotations. In Figure \ref{fig:sentence-stats}, we report the average length, number of frames and number of annotated FEs for each document. We observe, for instance, that the English and Brazilian Portuguese corpora contain, on average, twice as many annotated frames as the other languages, indicating a high density of annotations. On the other hand, the French dataset contains longer documents than the other languages but displays a similar number of frames, indicating a lower annotation density.

In the next section we further explore these annotation differences.

\subsection{Similarity of annotations}
\label{sec:content-composition}
\begin{figure}
    \centering
    \includegraphics[width=\linewidth]{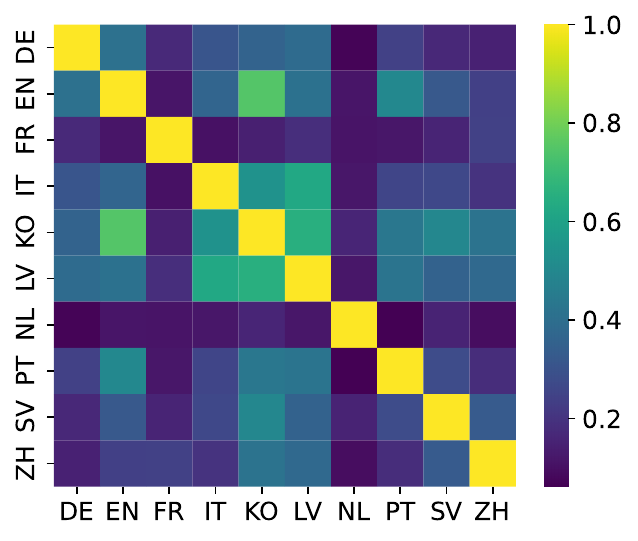}
    \caption{Cosine similarity between the centroids of each language computed using frame similarity.}
    \label{fig:similarity-centroids}
\end{figure}
In this section, we investigate in more detail how the ten resources converge and diverge with respect to their frame annotations, analyzing frame similarity. To do so, we rely on the FFICF measure, which adapts TF-IDF to derive typicality scores for each frame \citep{vossen-etal-2020-large}. The frame frequency is computed with respect to all the documents in the mFNC corpus. 
For each language we compute the centroid of its vectors and compare it with the centroids of other languages using cosine similarity. Hence, two languages are similar if they share a similar set of typical frames.

Results are shown in Figure \ref{fig:similarity-centroids}.

\paragraph{Topic Specificity}
We observe that Dutch and French have the most dissimilar annotations when compared to the BFN corpus and to the other resources. We speculate that this is due to the domain-specificity of annotated texts. For instance, the Dutch FrameNet corpus only collects documents that report on specific events (e.g., disease outbreak and wildfires) \citep{vossen-etal-2020-large}, resulting in \textsc{Catastrophe} being one of the most commonly annotated frames. Similarly, the French FrameNet corpus includes specialized documents in the medical and political domains \citep{djemaa2016asfalda,candito2014sequoia}. As a consequence, these resources show a biased distribution of frames, which reflects their topic-specificity. 

\paragraph{Annotation Practice}
In contrast, the set of most typical Korean frames is similar to that of the BFN corpus, which reflects the use of a projective annotation practice. A similar behavior can be observed in the Swedish corpus, which also partially relies on projective annotations.

\section{Experiments}
\label{sec:experiments}
In this section we demonstrate the effectiveness of the mFNC by training different FSP models on both the BFN corpus and on the mFNC, and compare their performance.

\subsection{Experimental Setting}
We experiment with two architectures representative of SotA approaches: the multi-stage approach proposed in LOME \citep{xia-etal-2021-lome}, where an XLM-RoBERTa encoder model is fine-tuned to parameterize a CRF layer used to extract spans from the input sentence that are then classified using two MLP layers (one for frames and one for FEs); and a generative approach inspired by \citet{kalyanpur2020seq2seqfsp}, where FSP is framed as a generative seq2seq task. In particular, we use the sentinel approach described in \citet{raman-etal-2022-transforming} and fine-tune the small and base versions of mT5 \citep{xue-etal-2021-mt5}.

We train LOME using the default hyperparameters defined by the authors on an RTX3090 with 24 GB of VRAM for a maximum of $50$ epochs, stopping the model if it does not improve its performance on the validation split for 3 consecutive epochs. We fine-tune the mT5 models using the hyperparameters suggested in \citet{raman-etal-2022-transforming} for $30$ epochs on an RTX6000 with 48 GB of VRAM, using the same early-stopping mechanism used for LOME.

\subsection{Results on the BFN corpus}
\begin{table}[t]
    \centering
    \begin{tabular}{cc}
        \toprule
        \textbf{Model} & \textbf{F1} $\uparrow$ \\ \midrule
        \citet{swayamdipta:17}$^*$ & 0.733 \\
        \citet{lin-etal-2021-graph}$^*$ & 0.763 \\
        \citet{devasier-etal-2024-robust}$^*$ & 0.775 \\ 
        mT5 small & 0.560 \\
        mT5 base & 0.583 \\
        LOME & 0.800 \\
        \midrule
        mT5 small$^\dagger$ & 0.677 \\ 
        mT5 base$^\dagger$ & 0.668 \\ 
        LOME$^\dagger$ & \textbf{0.812} \\ 
        \bottomrule
    \end{tabular}
    \caption{\textbf{Training on the mFNC outperforms training on the BFN on the English language.} Traditional micro-F1 score on the target identification and classification task. Results marked with $^*$ are taken from \citet{devasier-etal-2024-robust}. Results marked with $\dagger$ are trained on the mFNC.}
    \label{tab:frame-f1-on-bfn}
\end{table}

Before evaluating the performance of FSP systems in a multilingual setting, we evaluate whether training on the mFNC maintains competitive performance compared to training only on the BFN corpus. 

In Table \ref{tab:frame-f1-on-bfn} we report the traditional micro-averaged F1 score of the models trained on the BFN corpus and on the mFNC in the target classification task, i.e., on frame-evoking LU detection and frame attribution.

A prediction is considered correct when it fully matches the gold annotation. We compare our results with those from (i) \citet{swayamdipta:17}, who frame the task as a token classification problem relying on pre-trained static word embeddings and an LSTM model, (ii) \citet{lin-etal-2021-graph}, who frame the problem as a graph generation problem, fine-tuning a BERT-based encoder model, and (iii) \citet{devasier-etal-2024-robust}, who formulate the task as a QA task solved by fine-tuning a RoBERTa encoder model. 

We find that training on the mFNC maintains competitive performance with models trained on the BFN corpus, demonstrating that additional training data on other languages does not harm performance. Additionally, we demonstrate that training LOME on the mFNC outperforms existing SotA models. This result encourages novel FSP systems to be trained on the mFNC.

\subsection{Results on the mFNC}
\label{subsec:mfc-results}

\begin{table*}[ht]
    \centering
    \begin{tabular}{cccccc|ccc}
    \toprule
    & \multirow[c]{2}{*}{Model} & \multirow[c]{2}{*}{Train} & \multicolumn{3}{c}{Frame} & \multicolumn{3}{c}{FE} \\ \cmidrule{4-9}
    & & & Precision & Recall & F1 & Precision & Recall & F1 \\ \midrule
    
    & \multirow[c]{2}{*}{LOME} & BFN & 0.25 { \tiny $\pm$ 0.23 } & 0.60 { \tiny $\pm$ 0.22 } & 0.32 { \tiny $\pm$ 0.22 } & 0.15 { \tiny $\pm$ 0.18 } & 0.33 { \tiny $\pm$ 0.20 } & 0.19 { \tiny $\pm$ 0.18 } \\
    & & mFNC & \textbf{\underline{0.77}} { \tiny $\pm$ 0.11 } & \textbf{\underline{0.65}} { \tiny $\pm$ 0.20 } & \textbf{\underline{0.69}} { \tiny $\pm$ 0.15 } & \textbf{\underline{0.61}} { \tiny $\pm$ 0.15 } & \textbf{\underline{0.54}} { \tiny $\pm$ 0.18 } & \textbf{\underline{0.56}} { \tiny $\pm$ 0.15 } \\
    \cmidrule{1-9}
    & \multirow[c]{2}{*}{mT5} & BFN & 0.14 { \tiny $\pm$ 0.15 } & 0.43 { \tiny $\pm$ 0.19 } & 0.19 { \tiny $\pm$ 0.15 } & 0.08 { \tiny $\pm$ 0.10 } & 0.19 { \tiny $\pm$ 0.13 } & 0.10 { \tiny $\pm$ 0.10 } \\
    & & mFNC & \underline{0.53} { \tiny $\pm$ 0.12 } & \underline{0.57} { \tiny $\pm$ 0.13 } & \underline{0.55} { \tiny $\pm$ 0.12 } & \underline{0.40} { \tiny $\pm$ 0.13 } & \underline{0.41} { \tiny $\pm$ 0.13 } & \underline{0.40} { \tiny $\pm$ 0.13 } \\
    \cmidrule{1-9}
    & \multirow[c]{2}{*}{mT5 small} & BFN & 0.11 { \tiny $\pm$ 0.12 } & 0.37 { \tiny $\pm$ 0.17 } & 0.16 { \tiny $\pm$ 0.13 } & 0.06 { \tiny $\pm$ 0.08 } & 0.15 { \tiny $\pm$ 0.10 } & 0.08 { \tiny $\pm$ 0.09 } \\
    & & mFNC & \underline{0.59} { \tiny $\pm$ 0.14 } & \underline{0.60} { \tiny $\pm$ 0.14 } & \underline{0.59} { \tiny $\pm$ 0.14 } & \underline{0.42} { \tiny $\pm$ 0.13 } & \underline{0.42} { \tiny $\pm$ 0.14 } & \underline{0.42} { \tiny $\pm$ 0.13 } \\
    \bottomrule    
    \end{tabular}
    \caption{\textbf{Training on the mFNC outperforms training on the BFN on all metrics}. FairEval metrics computed on the mFNC averaged over the ten languages.}
    \label{tab:aggregated-results}
\end{table*}

In Table \ref{tab:aggregated-results} we report the performance of our models trained on the BFN corpus and on the mFNC and evaluated on the testing set of the mFNC by aggregating over the ten languages. Unlike Table \ref{tab:frame-f1-on-bfn}, we report micro-averaged precision, recall and F1 scores computed using the suggested configuration of the FairEval framework \citep{Ortmann-Faireval}. This allows us to account for correctly identified but mislabeled LUs and FEs, and for predictions whose boundaries partially overlap with the gold ones. 


Note that the metrics grouped under the \textit{Frame} column measure performance in identifying and labeling frame-evoking LUs. In turn, the \textit{FE} column evaluates the identification and classification of FEs.
In other words, the metrics grouped under the FE column evaluate the end-to-end performance of the FSP model, accounting for errors propagated from the frame identification process.

\paragraph{Multilingual performance}
The results provide strong evidence of the impact of the mFNC on the parsers' performance. Each model, regardless of the employed architecture, greatly outperforms the corresponding variant trained only on the English corpus. In particular, LOME trained on the mFNC consistently outperforms all the other FSP models across all the evaluated dimensions, providing evidence that FSP benefits from being treated as a sequence labeling task rather than as a seq2seq one.

\begin{figure}[t]
    \centering
    \includegraphics[width=\linewidth]{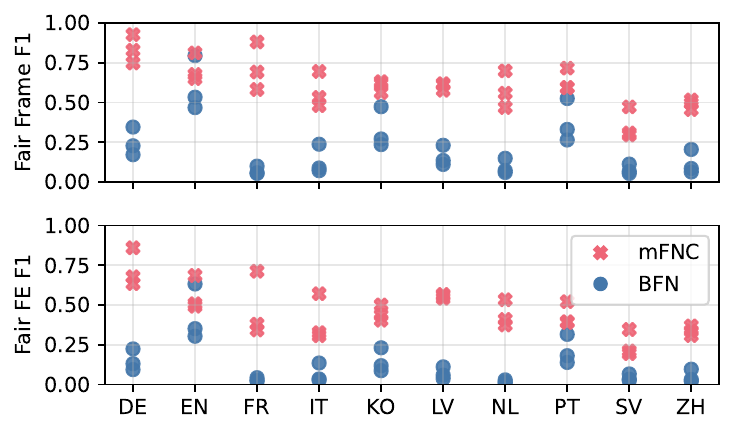}
    \caption{\textbf{Training on the mFNC outperforms training on the BFN on all languages.} F1 scores on frame and FEs performance on each language when trained on BFN vs mFNC.}
    \label{fig:language-wise-performances}
\end{figure}

Although training on the mFNC consistently improves results, performance gains are not equally distributed across the ten languages in the corpus. 

\paragraph{Language-wise improvements}
In Figure \ref{fig:language-wise-performances} we show how F1 scores vary across different languages when training on the BFN corpus or on the mFNC (see Table \ref{tab:results-by-language} in the Appendix for a more detailed overview). Similar to the results in Table \ref{tab:frame-f1-on-bfn}, performance on the English language is similar across the English-only and multilingual training settings, while it improves significantly on other languages, particularly in German, French, Dutch and Latvian. The Swedish corpus is the most challenging one, showing less pronounced improvements.

While we do not have a definitive explanation for this behavior, we found that the Swedish FrameNet has the largest coverage of annotated frames, annotating 78\% of BFN frames compared to 64\% for the BFN corpus. We speculate that the unbalanced nature of the mFNC (cf. Figure \ref{fig:splits-occurences}) might hamper generalization to infrequent frames, leaving open the question of whether it is possible to counter-balance this phenomenon during training (e.g., by annealing frequent frames), or pre-processing of the dataset (e.g., by performing data augmentation based on the hierarchical structure defined in the BFN). 

This result might also indicate that even when trained on multilingual data, the parser does not generalize to unseen languages. However, in the next section we demonstrate the opposite.

\subsubsection{Cross-lingual Generalization}
\label{subsec:ood-generalization}
\begin{table}[ht]An annotation is defined by the span that 283
activates a frame, the activated frame, and the list 284
of annotated arguments. Each argument is defined 285
by a role and the span that it classifies.
    \centering
    \begin{tabular}{ccccc}
    \toprule
    \multicolumn{2}{c}{Metric} & BFN & mFNC $\setminus$ SW & mFNC \\
    \midrule
    \multirow[c]{3}{*}{Frame} & P & 0.07 & \underline{0.30} & \textbf{0.55} \\
     & R & 0.26 & \underline{0.28} & \textbf{0.42} \\
     & F1 & 0.11 & \underline{0.29} & \textbf{0.47} \\
    \midrule
    \multirow[c]{3}{*}{FE} & P & 0.04 & \underline{0.20} & \textbf{0.36} \\
     & R & \underline{0.20} & \underline{0.20} & \textbf{0.33} \\
     & F1 & 0.07 & \underline{0.20} & \textbf{0.35} \\
    \bottomrule
    \end{tabular}
    \caption{\textbf{Training on the mFNC achieves better cross-lingual generalization than the BFN.} Performance of LOME trained on the BFN corpus, the mFNC without Swedish data (mFNC $\setminus$ SW) and on the full mFNC on the same setting as Table \ref{tab:aggregated-results}. Best results are in bold. Best between the mFNC $\setminus$ SW and the mFNC are underlined.}
    \label{tab:ood-evaluation}
\end{table}

In light of previous findings, we evaluate the impact that training on the mFNC has in cross-lingual settings by removing the Swedish dataset from the mFNC and training LOME from scratch on it. We follow the same experimental setting of previous experiments. Table \ref{tab:ood-evaluation} reports the results, showing that models trained on the mFNC have stronger cross-lingual generalization abilities compared to training on the BFN corpus. This further demonstrates the impact that the mFNC can have on future FSP models.

\section{Discussion}
\label{sec:discussion}
The results described in Section \ref{sec:experiments} demonstrate that current SotA FSP models trained on the BFN corpus struggle in cross-lingual settings. Their results, however, greatly improve when trained on our corpus, both in multilingual (Table \ref{tab:aggregated-results}) and cross-lingual (Table \ref{tab:ood-evaluation}) settings. 

Moreover, our results indicate that models tailored to the FSP task (LOME) perform significantly better than more general seq2seq approaches (mT5-based models). 

\paragraph{Integrating the mFNC with other resources}
As addressed in Section \ref{sec:dataset-composition} and in Figure \ref{fig:splits-occurences}, the mFNC is unbalanced with respect to the number of annotated frames. The effectiveness of FSP models is directly affected by this aspect, as illustrated in Section \ref{subsec:ood-generalization}. Although in this paper we only collect corpora annotated using the BFN, previous works (e.g., \citet{conia-etal-2022-semantic}) have shown that integrating additional semantic resources such as PropBank \citep{pradhan-etal-2022-propbank} and VerbNet \citep{palmer2017verbnet} results in better FSP performance. 

We refrained from adopting this approach because of the different nature of these resources compared to BFN\footnote{For example, PropBank frames are more focused on the lexical level than the semantic one, unlike BFN's frames \citep{bonial-etal-2014-propbank}.}. Nonetheless, it is worth investigating whether the mFNC could be extended by relying on recent efforts at aligning BFN with other resources \citep{lacalle2016predicatematrix}, which can result in broader linguistic coverage, for example by integrating the Polish \citep{jindal-etal-2022-universal} or Arabic \citep{palmer-etal-2008-pilot} PropBank-annotated datasets. 

\paragraph{Extending the mFNC}
The mFNC is the first corpus of multilingual documents annotated using BFN frames, but it is still characterized by limited coverage when compared to other multilingual datasets. For example, there is a lack of representation for Middle-Eastern or African languages. Possible approaches to overcome this limitation include translating texts and projecting their annotations \citep{yu-etal-2022-beyond}, as discussed in Section \ref{sec:data-selection}. We remark, however, that fully automating this approach might produce imprecise annotations that do not take into account the tight relationship between the lexical, semantic and cultural dimensions of language. Other promising approaches include relying on (L)LMs to generate annotated sentences by explicitly defining the semantics of a frame as found in the original BFN resource \citep{cui-swayamdipta-2024-annotating}. In this context, the mFNC can serve as a repository of multilingual examples that show the linguistic diversity spanned by a frame.

\paragraph{Frame-based Linguistic Analyses}
In this paper, we focused on the impact that the mFNC has on the training of FSP models. Nonetheless, the corpus's value extends beyond this application. By harmonizing ten language-specific datasets across typologically and conceptually diverse languages, the mFNC also enables a wide range of cross-lingual analyses at scale. These include investigating why semantically equivalent expressions evoke different frames across languages \citep{yong-etal-2022-frame}, how language-specific phenomena such as compound nouns are realized differently \citep{ponkiya-etal-2021-framenet}, and how conceptual metaphors differ across languages \citep{otmakhova-etal-2026-animals}.

\section{Conclusion}
\label{sec:conclusion}
This paper presented the mFNC, a multilingual dataset harmonizing ten language-specific resources annotated using FrameNet. This contribution addresses the lack of multilingual training and evaluation data for the Frame Semantic Parsing task, demonstrating that training on multilingual data substantially improves the performance of multilingual and cross-lingual systems. Beyond parsing, the mFNC allows researchers to explore new directions for comparative research on conceptual and frame-semantic differences across languages by relying on a large, unified resource.

\section{Limitations}
\label{sec:limitations}
In this section we discuss the main limitations of our contributions concerning two main aspects: the mFNC construction described in Section \ref{sec:dataset}, and the experiments presented in Section \ref{sec:experiments}.

\subsection{On constructing the mFNC}
\paragraph{Flattened linguistic diversity}
As shown in Figure \ref{fig:corpus_example}, Frame Semantic annotations are highly dependent on the grammatical and semantic patterns of each language. In Section \ref{sec:data-selection}, we combine language-specific corpora annotated using BFN, implicitly assuming that the conceptual structure of BFN correctly transfers to languages other than English. Nonetheless, we are aware that adopting this approach remains an open question in NLP and linguistic research.

Recent work demonstrates its limits \cite{ellsworth-etal-2021-framenet,hahm-etal-2020-crowdsourcing}, particularly when translations and projective annotations are used. These works, however, do not flag the assumption as incorrect. Rather, they argue that not all the frames defined in BFN apply equally to different languages, positing the existence of a language-agnostic portion of BFN \citep{culo2013englishgerman}. In this context, the mFNC can serve as a research tool to identify this subset using data-driven approaches \citep{baker-etal-2018-frame,baker-lorenzi-2020-exploring}.

\paragraph{Domain bias}
As highlighted in Section \ref{sec:content-composition}, the mFNC comprises some resources that are domain-specific. Specifically, the Dutch and French corpora differ from the other FrameNets in that they annotate documents that focus on specific topics. While this thematic diversity can be beneficial, it also induces a representational bias whereby some frames may be under-represented in a resource due to the predominant topics of its documents (see for instance the top 5 most common frames used by each corpus in Table \ref{app:tab:top-5} in the Appendix). Assessing the impact of this limitation and mitigating its effects is an important step toward ensuring more cross-lingual balance in the mFNC.

\paragraph{Annotation bias}
Related to the previous limitations, combining the corpora described in Section \ref{sec:data-selection} also assumes consistency across the human annotations of each resource. This assumption may introduce additional biases in the mFNC. For instance, \citet{dumitrache2018frameambiguityannotations} and \citet{hahm-etal-2020-crowdsourcing} observed cross-annotator differences in resources annotated using BFN, in both expert and crowd-sourced annotations.

Although this does not necessarily translate to low quality annotations, it might result in an uneven distribution of frames across corpora, similar to the domain bias described previously. Overcoming this limitation is not trivial, due to the inherent complexity of the annotation task. On the other hand, we argue that the mFNC can serve as a resource for analyzing annotation perspectives and differences \citep{cabitza2023perspectivist}, such as cross-language and cross-cultural ones.

We also note that retaining the original tokenization of documents, as described in Section \ref{sec:data-harmonization}, leads to combining different (possibly incompatible) tokenization practices \citep{habert1998towardstoken}. Similarly, recovering the full-text document through detokenization or tokenizing full-text documents using an automated tool may introduce noise, depending on the accuracy of the tool \citep{goot-2024-still}. Our approach is maximally conservative with respect to the design of each corpus, allowing researchers to study differences or apply refinements if needed. 

\paragraph{Language coverage}
The languages collected in the mFNC are mostly high-resource ones (e.g., English, German, French), which leaves open the question of how well the FSP systems evaluated in Section \ref{sec:experiments} generalize to low-resource languages. 
We already identified the extension of the mFNC to other languages as one of the main follow-ups (Section \ref{sec:discussion}) of this work. In this context, we highlight that there exist frame annotation efforts for low-resource languages (e.g., Arabic \citep{gargett-leung-2020-building} and Bengali \citep{datta2025bengalifn}), which represent promising integrations in this direction.

\subsection{On experimenting with the mFNC}
\paragraph{Assuming a correct frame}
As reported previously, different sets of frames might apply to the same sentence depending on how it is interpreted by the annotator \citep{dumitrache2018frameambiguityannotations,hahm-etal-2020-crowdsourcing}.
This identifies an important limitation of FSP models as well. From a general perspective, the models of Section \ref{sec:experiments} treat the FSP task in a discriminative fashion by assuming that an LU is classified by a single, correct frame (and similarly for FEs). 
It follows that those models inherit the (unknown) biases induced by the annotator's perspective. Ongoing research on how to tackle this limitation, which is shared by other common NLP tasks \citep{frenda2025perspectivesurvey}, can benefit from the mFNC as an additional corpus for experimentation.

\paragraph{Recall vs precision}
The limitation discussed above also raises the question of whether the evaluation metrics used in Section \ref{sec:experiments} over- or under- estimate the applicability of the tested FSP systems. For instance, a system that reaches a high precision at the expense of a low recall might reflect a conservative parsing behavior. This might be beneficial in some settings, but for many of the FSP applications reviewed in Section \ref{sec:rw}, a low volume of predictions translates to a limited amount of available information. 
By relying on the FairEval framework \citep{Ortmann-Faireval}, we partially address this problem, so that correctly identified but mislabeled spans are treated as \textit{half-correct} predictions. Nonetheless, evaluating whether the metrics of Section \ref{sec:experiments} improve performance on downstream FSP applications remains an open problem that requires further research.

\paragraph{Diverse morphologies}
The models evaluated in Section \ref{sec:experiments} rely on Transformer-based  multilingual models, which assumes that their internal representations act as a cross-lingual bridge. It is well known, however, that the tokenization phase of those models might disfavor some languages \citep{petrov2023tokenization}. Coupled with the limitations discussed previously, this poses further questions on the abilities of FSP models to process low-resource languages. Although out of scope for this paper, experimenting with backbone models that rely on a different tokenization strategy (e.g., ByT5 \citep{xue-etal-2022-byt5}) might result in better cross-lingual generalization.





\section*{Acknowledgments}
We wish to thank the anonymous reviewers and area chairs for their valuable comments; Ines Rehbein for providing us with access to the Salsa dataset; Arianna Graciotti for proofreading an early draft of the paper; Carmelo Caruso, Ludovica Pannitto and the \textit{Laboratorio Sperimentale} of the Department of Modern Languages, Literatures and Cultures (University of Bologna) for providing access to their computational resources.  
Beatrice Fiumanò and Valentina Presutti are supported by INFINITY: a EU Horizon Europe project under Grant Agreement No 101233051. Beatrice Fiumanò is funded by the National Recovery and Resilience Plan (NRRP), funded by the European Union – NextGenerationEU - Mission 4 "Education and Research", Component 1 "Enhancement of the offer of educational services: from nurseries to universities” - Investment 4.1 “Extension of the number of research doctorates and innovative doctorates for public administration and cultural heritage”.(DM 118/2023). 

\bibliography{references}

\appendix

\section{Appendix}
\label{sec:appendix}

\begin{figure}[ht!]
    \centering
    \includegraphics[width=\linewidth]{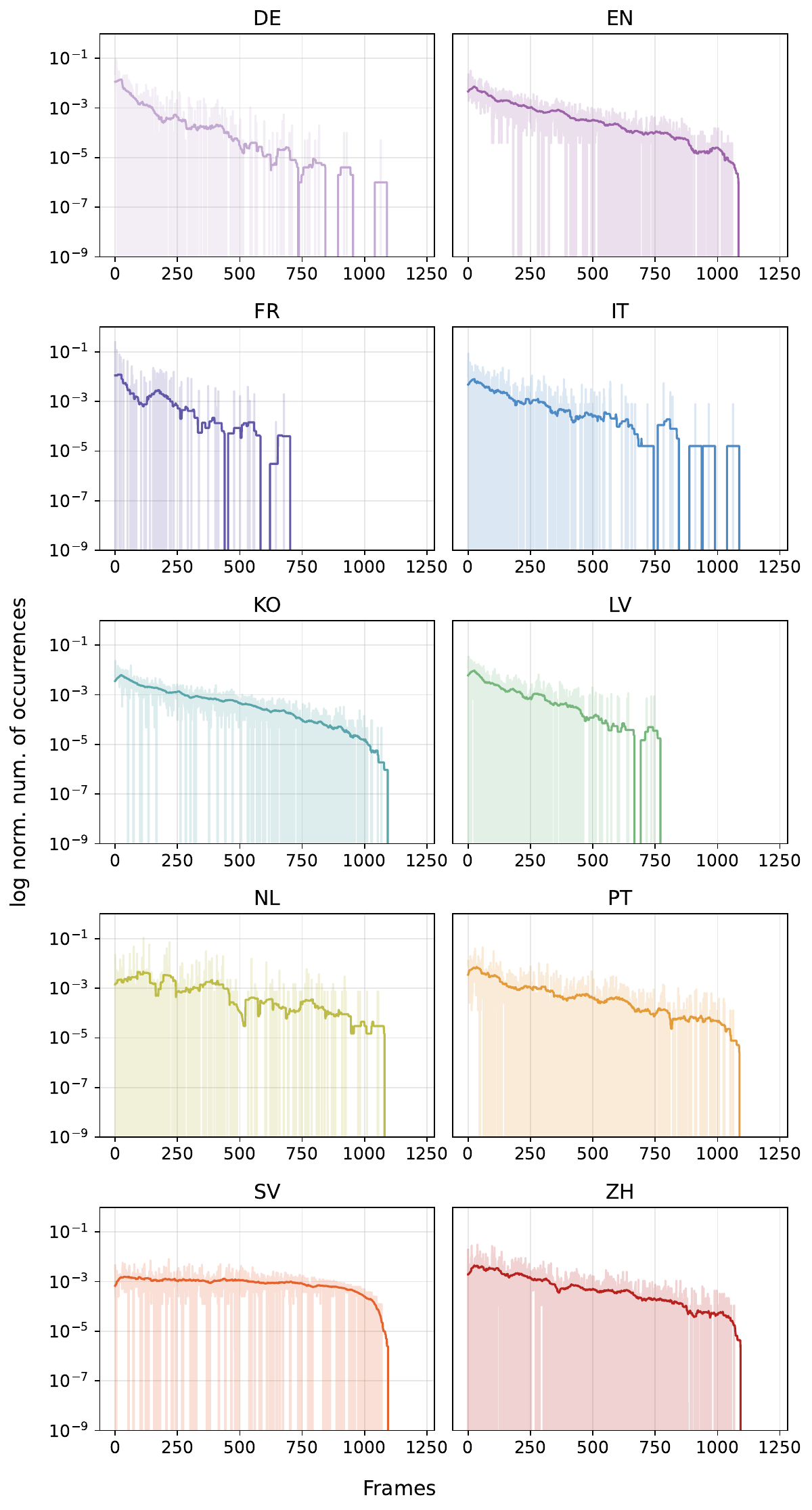}
    \caption{Number of occurrences of \textit{all} BFN frames (in log space) for each resource of Table \ref{tab:datasets}.}
    \label{app:fig:splits-occurrences-per-language}
\end{figure}


\begin{table*}
    \begin{subtable}[b]{0.5\linewidth}
    \centering
    \caption{DE}
        \begin{tabular}{rp{12em}}
        \toprule
        \# & Frame \\
        \midrule
        1852 & \textsc{Calendric\_unit} \\
        1623 & \textsc{Telling} \\
        1565 & \textsc{People} \\
        1368 & \textsc{Political\_locales} \\
        796 & \textsc{Request} \\
        \bottomrule
        \end{tabular}
    \end{subtable}%
    \begin{subtable}[b]{0.5\linewidth}
    \centering   
        \caption{EN}
        \begin{tabular}{rp{12em}}
        \toprule
        \# & Frame \\
        \midrule
        842 & \textsc{Weapon} \\
        672 & \textsc{Locale\_by\_use} \\
        603 & \textsc{Statement} \\
        556 & \textsc{Political\_locales} \\
        451 & \textsc{Leadership} \\
        \bottomrule
        \end{tabular}
    \end{subtable}%

    \begin{subtable}[b]{0.5\linewidth}
        \centering   
        \caption{FR}
        \begin{tabular}{rp{12em}}
            \toprule
            \# & Frame \\
            \midrule
            1599 & \textsc{Causation} \\
            786 & \textsc{Evidence} \\
            514 & \textsc{Commerce\_buy} \\
            414 & \textsc{Commerce\_sell} \\
            318 & \textsc{Reason} \\
            \bottomrule
        \end{tabular}
    \end{subtable}%
    \begin{subtable}[b]{0.5\linewidth}
        \centering   
        \caption{IT}
        \begin{tabular}{rp{12em}}
            \toprule
            \# & Frame \\
            \midrule
            109 & \textsc{Statement} \\
            47 & \textsc{Arriving} \\
            37 & \textsc{Attempt} \\
            33 & \textsc{Kinship} \\
            33 & \textsc{Desiring} \\
            \bottomrule
        \end{tabular}
    \end{subtable}%

    \begin{subtable}[b]{0.5\linewidth}
        \centering    
        \caption{KO}
        \begin{tabular}{rp{12em}}
        \toprule
        \# & Frame \\
        \midrule
        505 & \textsc{Statement} \\
        329 & \textsc{Experiencer\_focus} \\
        316 & \textsc{Locale\_by\_use} \\
        253 & \textsc{Leadership} \\
        248 & \textsc{Possession} \\
        \bottomrule
        \end{tabular}
    \end{subtable}%
    \begin{subtable}[b]{0.5\linewidth}
        \centering
        \caption{LV}
        \begin{tabular}{rp{12em}}
        \toprule
        \# & Frame \\
        \midrule
        475 & \textsc{Telling} \\
        466 & \textsc{Statement} \\
        349 & \textsc{Possession} \\
        346 & \textsc{Existence} \\
        343 & \textsc{Arriving} \\
        \bottomrule
        \end{tabular}
    \end{subtable}%

    \begin{subtable}[b]{0.5\linewidth}
        \centering
        \caption{NL}
        \begin{tabular}{rp{12em}}
        \toprule
        \# & Frame \\
        \midrule
        149 & \textsc{Catastrophe} \\
        97 & \textsc{Suspicion} \\
        77 & \textsc{Cause\_harm} \\
        65 & \textsc{Participation} \\
        55 & \textsc{Committing\_crime} \\
        \bottomrule
        \end{tabular}
    \end{subtable}%
    \begin{subtable}[b]{0.5\linewidth}
        \centering
        \caption{PT}
        \begin{tabular}{rp{12em}}
        \toprule
        \# & Frame \\
        \midrule
        354 & \textsc{Degree} \\
        311 & \textsc{Cardinal\_numbers} \\
        248 & \textsc{Negation} \\
        237 & \textsc{Locative\_relation} \\
        157 & \textsc{Possession} \\
        \bottomrule
        \end{tabular}
    \end{subtable}%

    \begin{subtable}[b]{0.5\linewidth}
        \centering       
        \caption{SV}
        \begin{tabular}{rp{12em}}
        \toprule
        \# & Frame \\
        \midrule
        66 & \textsc{Emptying} \\
        58 & \textsc{Make\_noise} \\
        47 & \textsc{Self\_motion} \\
        42 & \textsc{Experiencer\_obj} \\
        42 & \textsc{Placing} \\
        \bottomrule
        \end{tabular}
    \end{subtable}%
    \begin{subtable}[b]{0.5\linewidth}
        \centering     
        \caption{ZH}
        \begin{tabular}{rp{12em}}
        \toprule
        \# & Frame \\
        \midrule
        288 & \textsc{Cause\_to\_make\_progress} \\
        261 & \textsc{Change\_position\_on\_a\_scale} \\
        249 & \textsc{Being\_in\_category} \\
        218 & \textsc{Amounting\_to} \\
        178 & \textsc{Causation} \\
    \bottomrule
    \end{tabular}
    \end{subtable}%
    \caption{Top 5 most common annotated frames for each of the language-specific corpora of Table \ref{tab:datasets}.}
    \label{app:tab:top-5}
\end{table*}

\begin{table*}[ht!]
    \scriptsize
    \centering
    \begin{tabular}{llllllllll}
    \toprule
     &  &  &  & \multicolumn{3}{c}{Frame} & \multicolumn{3}{c}{FE} \\
     &  &  &  & P & R & F1 & P & R & F1 \\
    lang & flag & model & dataset &  &  &  &  &  &  \\
    \midrule
    \multirow[t]{6}{*}{DE} & \multirow[t]{6}{*}{\emoji{emoji-imgs/de.png}} & \multirow[t]{2}{*}{LOME} & BFN & 0.217 & 0.853 & 0.346 & 0.139 & 0.587 & 0.224 \\
     &  &  & mFNC & \textbf{0.925} & \textbf{0.929} & \textbf{0.927} & \textbf{0.859} & \textbf{0.859} & \textbf{0.859} \\
    \cmidrule{3-10}
     &  & \multirow[t]{2}{*}{mT5} & BFN & 0.136 & 0.706 & 0.228 & 0.079 & 0.382 & 0.131 \\
     &  &  & mFNC & \textbf{0.748} & \textbf{0.749} & \textbf{0.748} & \textbf{0.632} & \textbf{0.636} & \textbf{0.634} \\
    \cmidrule{3-10}
     &  & \multirow[t]{2}{*}{mT5 small} & BFN & 0.101 & 0.583 & 0.172 & 0.057 & 0.268 & 0.094 \\
     &  &  & mFNC & \textbf{0.833} & \textbf{0.823} & \textbf{0.828} & \textbf{0.678} & \textbf{0.676} & \textbf{0.677} \\
    \midrule
    
    \multirow[t]{6}{*}{EN} & \multirow[t]{6}{*}{\emoji{emoji-imgs/us.png}} & \multirow[t]{2}{*}{LOME} & BFN & 0.744 & \textbf{0.852} & 0.794 & 0.595 & 0.673 & 0.632 \\
     &  &  & mFNC & \textbf{0.785} & 0.840 & \textbf{0.812} & \textbf{0.666} & \textbf{0.705} & \textbf{0.685} \\
    \cmidrule{3-10}
     &  & \multirow[t]{2}{*}{mT5} & BFN & 0.466 & 0.623 & 0.533 & 0.320 & 0.390 & 0.352 \\
     &  &  & mFNC & \textbf{0.614} & \textbf{0.691} & \textbf{0.650} & \textbf{0.491} & \textbf{0.494} & \textbf{0.492} \\
    \cmidrule{3-10}
     &  & \multirow[t]{2}{*}{mT5 small} & BFN & 0.382 & 0.603 & 0.468 & 0.267 & 0.351 & 0.304 \\
     &  &  & mFNC & \textbf{0.642} & \textbf{0.713} & \textbf{0.676} & \textbf{0.502} & \textbf{0.515} & \textbf{0.508} \\
    \midrule
    
    \multirow[t]{6}{*}{FR} & \multirow[t]{6}{*}{\emoji{emoji-imgs/fr.png}} & \multirow[t]{2}{*}{LOME} & BFN & 0.059 & 0.332 & 0.100 & 0.027 & 0.107 & 0.043 \\
     &  &  & mFNC & \textbf{0.874} & \textbf{0.887} & \textbf{0.880} & \textbf{0.712} & \textbf{0.713} & \textbf{0.712} \\
    \cmidrule{3-10}
     &  & \multirow[t]{2}{*}{mT5} & BFN & 0.031 & 0.225 & 0.054 & 0.015 & 0.061 & 0.024 \\
     &  &  & mFNC & \textbf{0.580} & \textbf{0.584} & \textbf{0.582} & \textbf{0.344} & \textbf{0.342} & \textbf{0.343} \\
    \cmidrule{3-10}
     &  & \multirow[t]{2}{*}{mT5 small} & BFN & 0.033 & 0.261 & 0.058 & 0.016 & 0.058 & 0.025 \\
     &  &  & mFNC & \textbf{0.704} & \textbf{0.679} & \textbf{0.691} & \textbf{0.382} & \textbf{0.378} & \textbf{0.380} \\
    \midrule
    
    \multirow[t]{6}{*}{IT} & \multirow[t]{6}{*}{\emoji{emoji-imgs/it.png}} & \multirow[t]{2}{*}{LOME} & BFN & 0.142 & \textbf{0.747} & 0.238 & 0.084 & 0.348 & 0.136 \\
     &  &  & mFNC & \textbf{0.700} & 0.690 & \textbf{0.695} & \textbf{0.564} & \textbf{0.578} & \textbf{0.571} \\
    \cmidrule{3-10}
     &  & \multirow[t]{2}{*}{mT5} & BFN & 0.051 & 0.341 & 0.088 & 0.022 & 0.090 & 0.035 \\
     &  &  & mFNC & \textbf{0.438} & \textbf{0.533} & \textbf{0.481} & \textbf{0.296} & \textbf{0.318} & \textbf{0.307} \\
    \cmidrule{3-10}
     &  & \multirow[t]{2}{*}{mT5 small} & BFN & 0.041 & 0.272 & 0.071 & 0.018 & 0.069 & 0.029 \\
     &  &  & mFNC & \textbf{0.530} & \textbf{0.534} & \textbf{0.532} & \textbf{0.330} & \textbf{0.323} & \textbf{0.327} \\
    \midrule
    
    \multirow[t]{6}{*}{KO} & \multirow[t]{6}{*}{\emoji{emoji-imgs/kr.png}} & \multirow[t]{2}{*}{LOME} & BFN & 0.364 & \textbf{0.679} & 0.474 & 0.187 & 0.303 & 0.231 \\
     &  &  & mFNC & \textbf{0.791} & 0.525 & \textbf{0.631} & \textbf{0.534} & \textbf{0.471} & \textbf{0.501} \\
    \cmidrule{3-10}
     &  & \multirow[t]{2}{*}{mT5} & BFN & 0.176 & 0.588 & 0.271 & 0.079 & 0.238 & 0.119 \\
     &  &  & mFNC & \textbf{0.495} & \textbf{0.651} & \textbf{0.562} & \textbf{0.378} & \textbf{0.436} & \textbf{0.405} \\
    \cmidrule{3-10}
     &  & \multirow[t]{2}{*}{mT5 small} & BFN & 0.155 & 0.487 & 0.236 & 0.059 & 0.166 & 0.087 \\
     &  &  & mFNC & \textbf{0.585} & \textbf{0.654} & \textbf{0.617} & \textbf{0.435} & \textbf{0.437} & \textbf{0.436} \\
    \midrule
    
    \multirow[t]{6}{*}{LV} & \multirow[t]{6}{*}{\emoji{emoji-imgs/lv.png}} & \multirow[t]{2}{*}{LOME} & BFN & 0.135 & \textbf{0.806} & 0.231 & 0.063 & \textbf{0.463} & 0.111 \\
     &  &  & mFNC & \textbf{0.855} & 0.435 & \textbf{0.577} & \textbf{0.797} & 0.410 & \textbf{0.542} \\
    \cmidrule{3-10}
     &  & \multirow[t]{2}{*}{mT5} & BFN & 0.076 & \textbf{0.649} & 0.136 & 0.033 & 0.310 & 0.059 \\
     &  &  & mFNC & \textbf{0.612} & 0.612 & \textbf{0.612} & \textbf{0.566} & \textbf{0.567} & \textbf{0.566} \\
    \cmidrule{3-10}
     &  & \multirow[t]{2}{*}{mT5 small} & BFN & 0.061 & 0.560 & 0.111 & 0.021 & 0.206 & 0.037 \\
     &  &  & mFNC & \textbf{0.619} & \textbf{0.619} & \textbf{0.619} & \textbf{0.555} & \textbf{0.563} & \textbf{0.559} \\
    \midrule
    
    \multirow[t]{6}{*}{NL} & \multirow[t]{6}{*}{\emoji{emoji-imgs/nl.png}} & \multirow[t]{2}{*}{LOME} & BFN & 0.088 & 0.491 & 0.149 & 0.017 & 0.127 & 0.030 \\
     &  &  & mFNC & \textbf{0.752} & \textbf{0.652} & \textbf{0.699} & \textbf{0.617} & \textbf{0.468} & \textbf{0.533} \\
    \cmidrule{3-10}
     &  & \multirow[t]{2}{*}{mT5} & BFN & 0.042 & 0.290 & 0.073 & 0.008 & 0.066 & 0.014 \\
     &  &  & mFNC & \textbf{0.446} & \textbf{0.495} & \textbf{0.469} & \textbf{0.373} & \textbf{0.375} & \textbf{0.374} \\
    \cmidrule{3-10}
     &  & \multirow[t]{2}{*}{mT5 small} & BFN & 0.034 & 0.260 & 0.060 & 0.007 & 0.058 & 0.012 \\
     &  &  & mFNC & \textbf{0.570} & \textbf{0.547} & \textbf{0.559} & \textbf{0.430} & \textbf{0.390} & \textbf{0.409} \\
    \midrule
    
    \multirow[t]{6}{*}{PT} & \multirow[t]{6}{*}{\emoji{emoji-imgs/br.png}} & \multirow[t]{2}{*}{LOME} & BFN & 0.515 & 0.536 & 0.525 & 0.311 & 0.320 & 0.316 \\
     &  &  & mFNC & \textbf{0.688} & \textbf{0.746} & \textbf{0.716} & \textbf{0.517} & \textbf{0.525} & \textbf{0.521} \\
    \cmidrule{3-10}
     &  & \multirow[t]{2}{*}{mT5} & BFN & 0.326 & 0.336 & 0.331 & 0.174 & 0.189 & 0.182 \\
     &  &  & mFNC & \textbf{0.574} & \textbf{0.618} & \textbf{0.595} & \textbf{0.397} & \textbf{0.394} & \textbf{0.396} \\
    \cmidrule{3-10}
     &  & \multirow[t]{2}{*}{mT5 small} & BFN & 0.262 & 0.267 & 0.264 & 0.131 & 0.148 & 0.139 \\
     &  &  & mFNC & \textbf{0.561} & \textbf{0.626} & \textbf{0.592} & \textbf{0.388} & \textbf{0.393} & \textbf{0.390} \\
    \midrule
    
    \multirow[t]{6}{*}{SV} & \multirow[t]{6}{*}{\emoji{emoji-imgs/se.png}} & \multirow[t]{2}{*}{LOME} & BFN & 0.072 & 0.260 & 0.113 & 0.040 & 0.202 & 0.067 \\
     &  &  & mFNC & \textbf{0.545} & \textbf{0.417} & \textbf{0.472} & \textbf{0.362} & \textbf{0.332} & \textbf{0.346} \\
    \cmidrule{3-10}
     &  & \multirow[t]{2}{*}{mT5} & BFN & 0.041 & 0.191 & 0.067 & 0.020 & 0.114 & 0.034 \\
     &  &  & mFNC & \textbf{0.318} & \textbf{0.301} & \textbf{0.309} & \textbf{0.211} & \textbf{0.211} & \textbf{0.211} \\
    \cmidrule{3-10}
     &  & \multirow[t]{2}{*}{mT5 small} & BFN & 0.033 & 0.164 & 0.054 & 0.015 & 0.092 & 0.026 \\
     &  &  & mFNC & \textbf{0.304} & \textbf{0.288} & \textbf{0.296} & \textbf{0.194} & \textbf{0.195} & \textbf{0.194} \\
    \midrule
    \multirow[t]{6}{*}{ZH} & \multirow[t]{6}{*}{\emoji{emoji-imgs/cn.png}} & \multirow[t]{2}{*}{LOME} & BFN & 0.135 & \textbf{0.428} & 0.205 & 0.075 & 0.135 & 0.097 \\
     &  &  & mFNC & \textbf{0.747} & 0.369 & \textbf{0.494} & \textbf{0.472} & \textbf{0.303} & \textbf{0.369} \\
    \cmidrule{3-10}
     &  & \multirow[t]{2}{*}{mT5} & BFN & 0.049 & 0.317 & 0.086 & 0.020 & 0.081 & 0.032 \\
     &  &  & mFNC & \textbf{0.440} & \textbf{0.471} & \textbf{0.455} & \textbf{0.306} & \textbf{0.310} & \textbf{0.308} \\
    \cmidrule{3-10}
     &  & \multirow[t]{2}{*}{mT5 small} & BFN & 0.037 & 0.239 & 0.063 & 0.013 & 0.051 & 0.021 \\
     &  &  & mFNC & \textbf{0.517} & \textbf{0.514} & \textbf{0.516} & \textbf{0.336} & \textbf{0.338} & \textbf{0.337} \\
    
    \bottomrule
    \end{tabular}
    \caption{Results obtained by each model on each language, computed using the FairEval framework.}
    \label{tab:results-by-language}
\end{table*}

\end{document}